\documentclass[10pt,twocolumn,letterpaper]{article}

\usepackage[pagenumbers]{wacv} 

\usepackage[utf8]{inputenc} 
\usepackage[T1]{fontenc}    
\usepackage{url}            
\usepackage{booktabs}       
\usepackage{amsfonts}       
\usepackage{nicefrac}       
\usepackage{microtype}      
\usepackage{mathtools}
\usepackage{adjustbox} 

\usepackage{subcaption}
\usepackage{graphicx}
\usepackage{tabularx}
\usepackage{array}
\usepackage{placeins}   
\usepackage{listings}
\usepackage{booktabs}     
\usepackage{tabularx}     
\usepackage{threeparttable} 
\usepackage{multirow}
\usepackage{relsize}
\usepackage{tikz}
\usepackage{enumitem}
\usepackage{siunitx}
\usetikzlibrary{arrows.meta, positioning}

\newcolumntype{Y}{>{\centering\arraybackslash}X} 

\definecolor{darkred}{RGB}{180,0,0}     
\definecolor{darkgreen}{RGB}{0,120,0}   

\definecolor{wacvblue}{rgb}{0.21,0.49,0.74}
\usepackage[pagebackref,breaklinks,colorlinks,allcolors=wacvblue]{hyperref}

\def\wacvPaperID{2675} 
\def\confName{WACV}
\def\confYear{2026}

\title{PoseBench3D: A Cross-Dataset Analysis Framework for 3D Human Pose Estimation via Pose Lifting Networks}

\author{%
\text{Saad Manzur}\thanks{Equal Contribution}\, \thanks{Corresponding Author}\; ,\;\quad
\text{Bryan Vela}\footnotemark[1]\; ,\;\quad
\text{Brandon Vela}\footnotemark[1]\; ,\\
\text{Aditya Agrawal},\quad
\text{Lan-Anh Dang-Vu},\quad
\text{David Li},\quad
\text{Wayne Hayes}\\
University of California, Irvine\\
{\tt\small \{smanzur, bjvela, bovela, agrawaa7, ldangvu, dli30, whayes\}@uci.edu}
}

\begin{document}
\maketitle

\begin{abstract}
Reliable three-dimensional human pose estimation (3D HPE) remains challenging due to the differences in viewpoints, environments, and camera conventions among datasets. As a result, methods that achieve near-optimal in-dataset accuracy often degrade on unseen datasets. In practice, however, systems must adapt to diverse viewpoints, environments, and camera setups—conditions that differ significantly from those encountered during training, which is often the case in real-world scenarios. Measuring cross-dataset performance is a vital process, but extremely labor-intensive when done manually for human pose estimation. To address these challenges, we automate this evaluation using PoseBench3D, a standardized testing framework that enables consistent and fair cross-dataset comparisons on previously unseen data. PoseBench3D streamlines testing across four widely used 3D HPE datasets via a single, configurable interface. Using this framework, we re-evaluate 18 methods and report over 100 cross-dataset results under Protocol 1: MPJPE and Protocol 2: PA-MPJPE, revealing systematic generalization gaps and the impact of common preprocessing and dataset setup choices. 
The PoseBench3D code is found at: \textcolor{magenta}{\href{https://github.com/bryanjvela/PoseBench3D}{https://github.com/bryanjvela/PoseBench3D}}.
\end{abstract}

\section{Introduction}
\label{sec:intro}
Three-dimensional Human Pose Estimation (HPE) has gathered substantial interest for its critical role in applications such as healthcare \citep{medicine1, medicine2}, action recognition \citep{actionrecognition1, actionrecognition2}, military operations \citep{military1, military2}, human-computer interaction \citep{robotinteraction1, robotinteraction2}, and virtual/augmented reality \citep{virtualreality1}—among many others. Despite remarkable progress in recent years \citep{liu2025tcpformer, tang20233d, motionbert2022, motionagformer2024, hu2021conditional, wang2020motion}, most work focuses on performance within a single, controlled dataset. The majority of existing work adopts either a direct-from-image approach or a two-stage approach, with the latter being widely used for its flexibility. In the two-stage approach, the first stage detects 2D keypoints, and the second stage, known as the ``lifting network'' \cite{2d3d_martinez2017simple,2d3d_gcn_zhao2019semantic,2d3d_gcn_liu2020comprehensive,2d3d_gcn_ci2019optimizing,2d3d_gcn_zou2021modulated,2d3d_hyp_li2019generating,2d3d_gcn_zeng2020srnet} maps the 2D keypoints to 3D pose estimations. However, these two-stage models show poor generalization across various different datasets \cite{i3d_cd_wang2020predicting,manzur2024human}.

\begin{figure}[!htb]
    \centering
    \begin{subfigure}{0.48\linewidth}
        \centering
        \includegraphics[width=0.5\columnwidth]{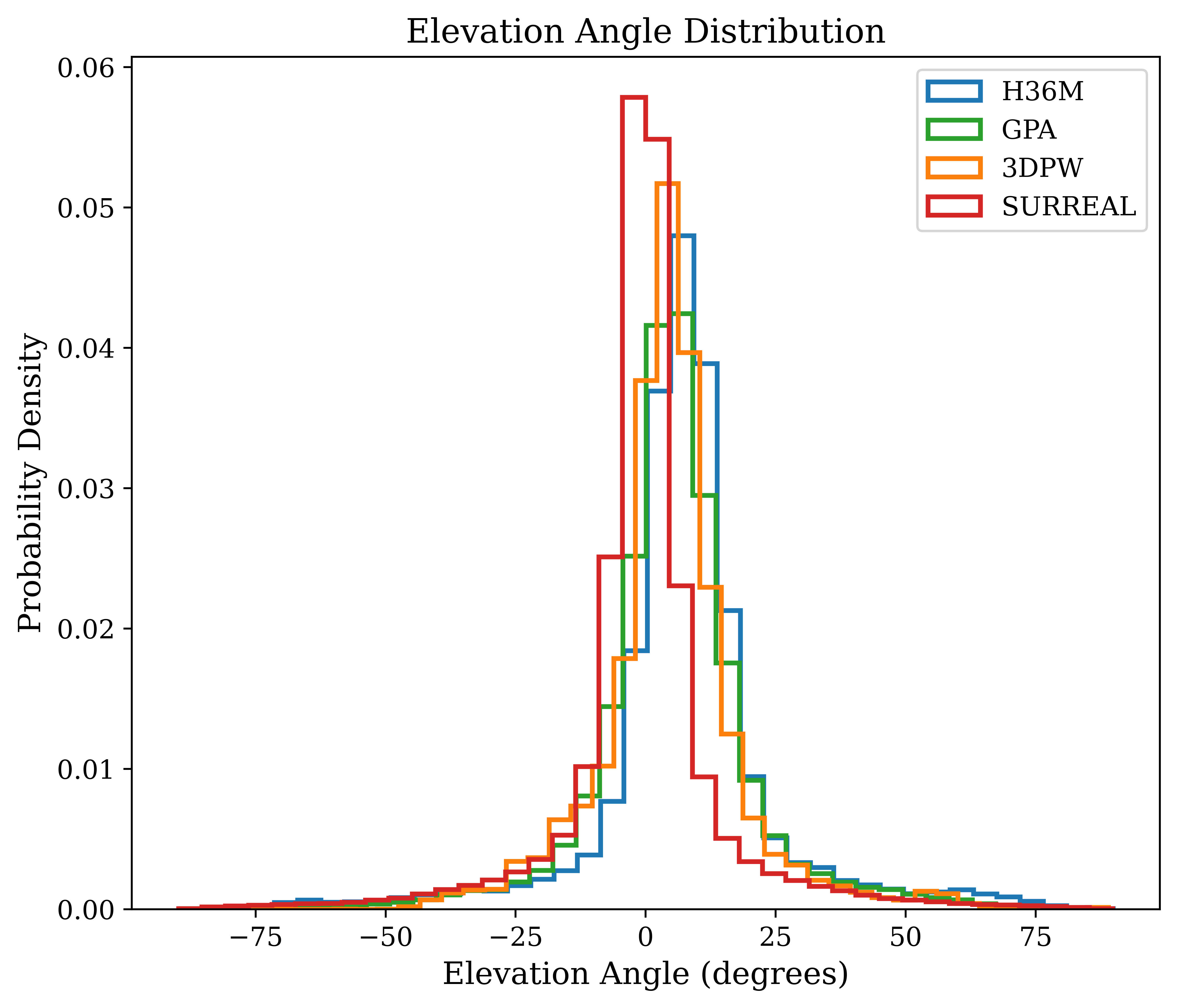}
        \caption{Elevation}
        \label{fig:elevation}
    \end{subfigure}
    \begin{subfigure}{0.48\linewidth}
        \centering
        \includegraphics[width=0.5\columnwidth]{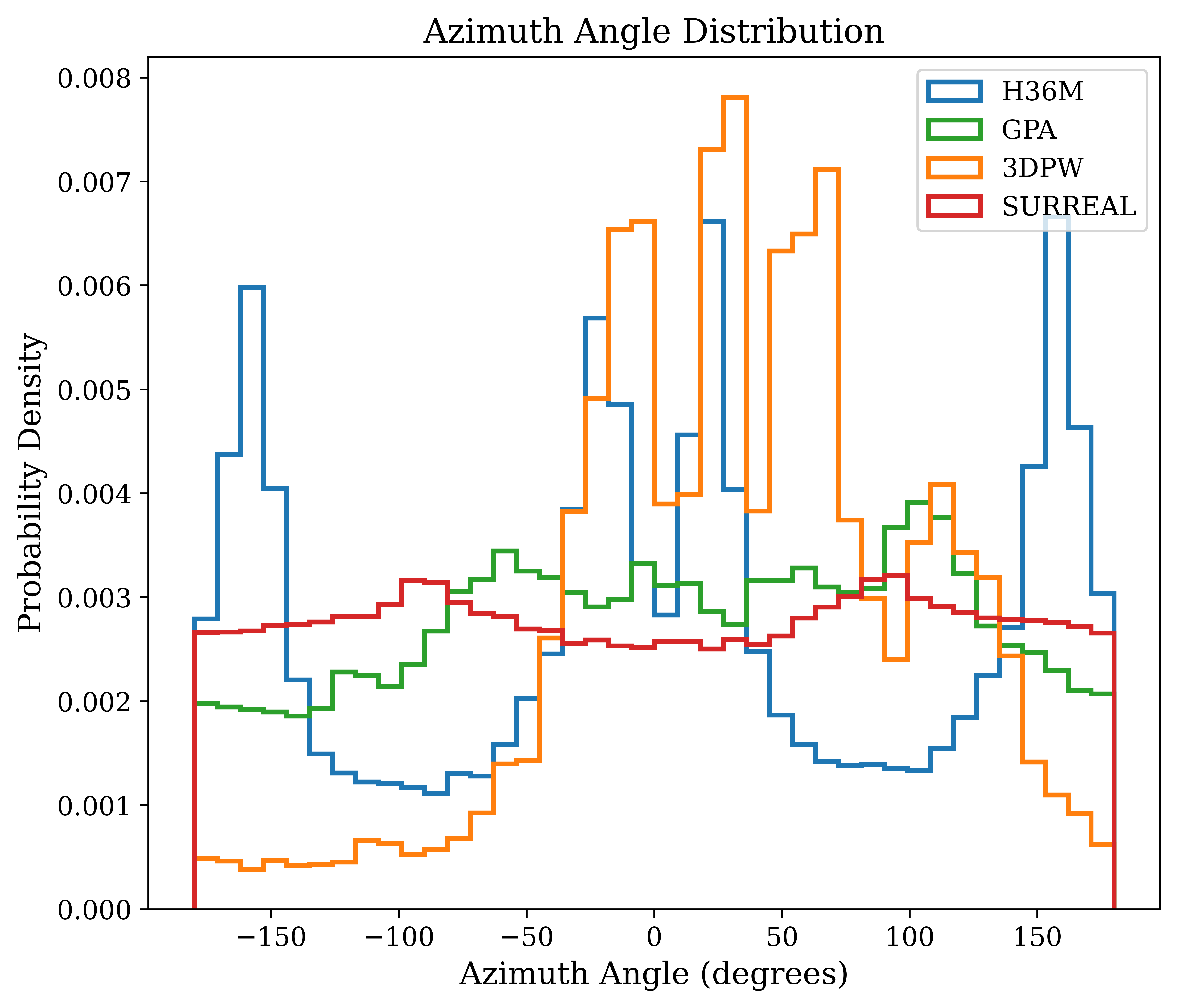}
        \caption{Azimuth}
        \label{fig:azimuth}
    \end{subfigure}
    \caption{Elevation and azimuth distribution across four datasets -- H36M, GPA, 3DPW, SURREAL. Elevation and azimuth are calculated relative to the subject's local axis (a) The elevation distribution. Notice how all the datasets have similar elevation profiles. (b) The azimuth distribution for the same configuration. We see stark differences across all datasets.}
    \label{fig:elev_and_azim}
\end{figure}

To assess the generalization of these models, we consider cross-dataset performance: the task of training on one dataset and testing on another. Machine learning models are generally susceptible to overfitting to the bias and variance of the data distribution they were trained on; this is particularly pronounced in 3D human pose estimation datasets due to variations in data collection procedures, where the underlying distributions of datasets often differ significantly. For example, the H36M dataset \citep{h36m} was captured using a four-camera setup, while GPA \citep{gpa} used five, and both 3DPW \citep{3dpw} and SURREAL \citep{varol17_surreal} relied on a single-camera configuration. In \cref{fig:elev_and_azim}, we show the frequency histogram of camera positions relative to the subject in spherical coordinates. While elevation distributions are largely consistent across datasets (\cf \cref{fig:elevation}), the azimuthal angles vary significantly (\cf \cref{fig:azimuth}).

\begin{figure}[!htb]
    \centering
    \begin{subfigure}{0.22\linewidth}
        \centering
        \includegraphics[width=0.95\columnwidth]{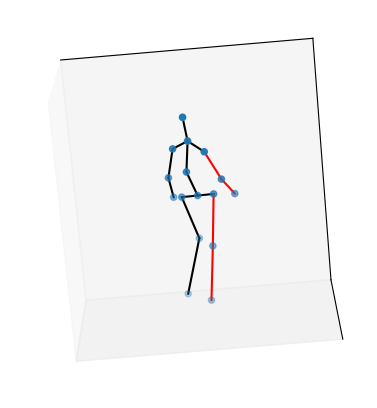}
        \caption{H36M}
        \label{fig:h36m_sample}
    \end{subfigure}
    \begin{subfigure}{0.22\linewidth}
        \centering
        \includegraphics[width=0.95\columnwidth]{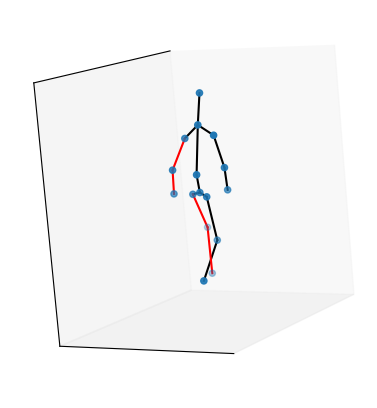}
        \caption{GPA}
        \label{fig:gpa_sample}
    \end{subfigure}
    \begin{subfigure}{0.22\linewidth}
        \centering
        \includegraphics[width=0.95\columnwidth]{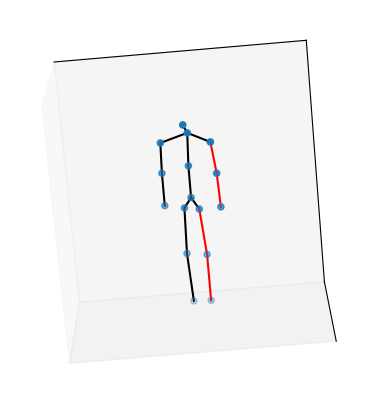}
        \caption{3DPW}
        \label{fig:3dpw_sample}
    \end{subfigure}
    \begin{subfigure}{0.22\linewidth}
        \centering
        \includegraphics[width=0.95\columnwidth]{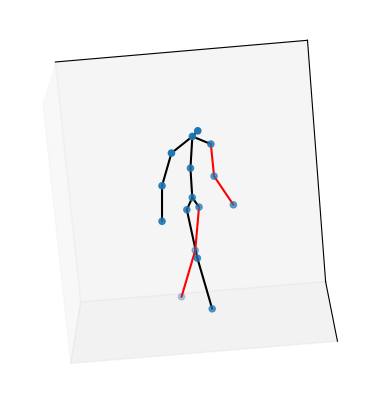}
        \caption{SURREAL}
        \label{fig:surreal_sample}
    \end{subfigure}
    \caption{Sample pose data points from H36M, GPA, 3DPW, and SURREAL. Compared to 3DPW, H36M shows a wider hip joint spread and more upright head posture.}
    \label{fig:dataset_samples}
\end{figure}

\begin{figure}[!htb]
    \centering
    \begin{subfigure}{0.48\linewidth}
        \centering
        \includegraphics[width=\linewidth]{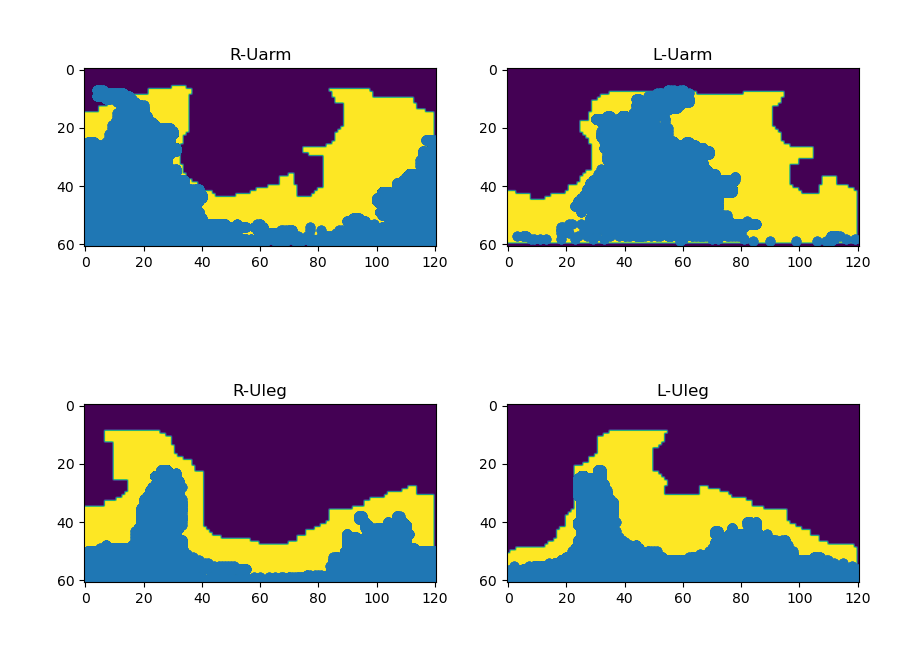}
        \caption{H36M}
        \label{fig:h36m_mobility_arm_leg}
    \end{subfigure}
    \hfill
    \begin{subfigure}{0.48\linewidth}
        \centering
        \includegraphics[width=\linewidth]{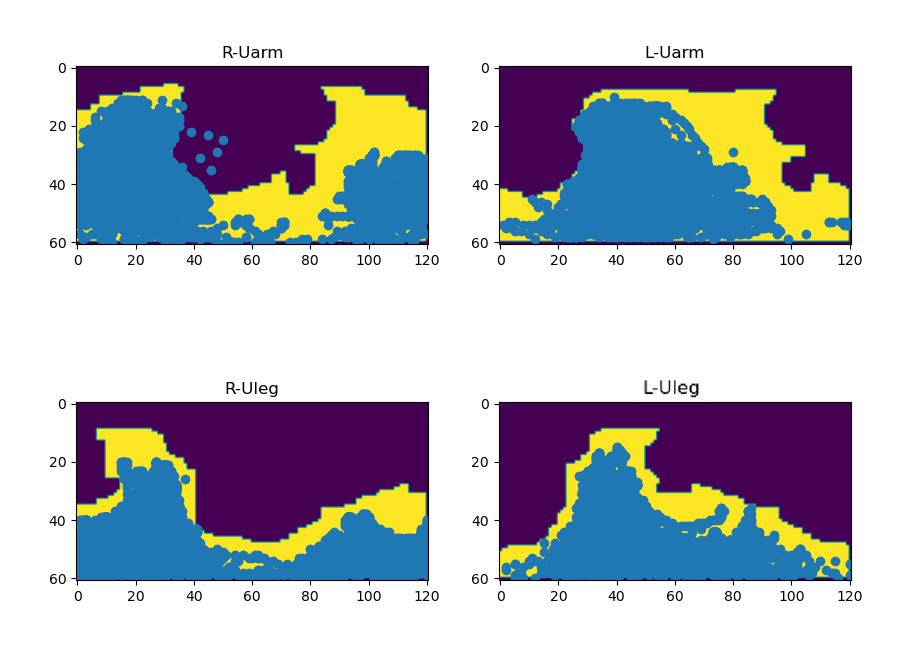}
        \caption{3DPW}
        \label{fig:3dpw_mobility_arm_leg}
    \end{subfigure}
    
    \caption{Scatter plots showing limb range of motion in H36M and 3DPW datasets. Blue represents regions of motion observed in the datasets, yellow represents the full range of motion based on physical limitations of limbs, and purple represents regions that cannot be reached without physical injury to the limb. Note the distribution differences between datasets.}
    \label{fig:mobility}
\end{figure}

Beyond camera setup, joint placement conventions also differ between datasets. For instance, there is currently no standardized convention for placing the hip or spine joints (\cf \cref{fig:dataset_samples})---note the difference in width of hips in \cref{fig:h36m_sample} and \cref{fig:3dpw_sample}. Additionally, since datasets are collected with varying action sets, the captured range of motion also varies. In \cref{fig:mobility}, the blue dots represent the recorded range of motion, whereas the yellow region shows the valid range. Since these two datasets do not explore the entire valid range of motion, normalization performed with one will not translate well onto the other.

\textbf{Research Gap.} These differences mean that, although any method may achieve near-optimal performance within its own dataset, it is likely to perform poorly across differing datasets. Measuring cross-dataset performance is paramount, but is extremely labor-intensive to perform manually. Our goal in this paper is to automate the task of cross-dataset comparison.

\textbf{Proposed Work.} To advance the field and encourage broader generalization, methods and architectures must be evaluated on multiple datasets under a consistent evaluation protocol---in order to better facilitate and automate cross-dataset evaluation. In this spirit, we introduce a unified test bench that consolidates several 3D human pose datasets within a standardized testing environment. This approach enables detailed analysis of generalization capabilities, method scalability, and performance trends across diverse environmental conditions—mirroring realistic scenarios in which settings are rarely uniform. Our aim is to narrow the existing research gap by offering insights into pose estimation that are simply not possible when relying solely on single-dataset evaluations.

Our \textbf{contributions} are summarized as follows:
\begin{enumerate}[
    leftmargin=0pt,        
    label=\textbf{\arabic*.},  
    itemsep=0pt,           
    wide=0pt               
]
    \item \textbf{Standardized Benchmarking Environment.} We introduce a standardized evaluation framework, PoseBench3D, that consolidates four commonly used human pose estimation datasets---H36M\cite{h36m}, GPA\cite{gpa}, 3DPW\cite{3dpw}, and SURREAL\cite{varol17_surreal}---while providing modularity for future datasets. Our framework simplifies cross-dataset evaluation by requiring, simply, a single user-supplied configuration file specifying model, experiment, and user customization details.

    \item \textbf{Re-Evaluation of 18 Methods.} We release detailed cross-dataset evaluation metrics for 18 established methods, generating over 100 previously unreported comparisons—shedding light on the need for standardized, comprehensive benchmarking in 3D human pose estimation.

    \item \textbf{Open and Extensible Design.} PoseBench3D is structured to accommodate both new models and datasets as the field progresses, encouraging fair, reproducible, and easily customizable evaluations through our open-source framework. We make our code publicly available.

    \item \textbf{Detailed Analysis.} We also investigate the impact of factors such as viewpoint distribution and Z-score standardization on a model's ability to generalize across datasets.
\end{enumerate}
\section{Related Work}

\textbf{Current Cross‑Dataset Evaluation.}
While several methods have attempted to address generalization, none have demonstrated success beyond testing on a single additional dataset beyond their training set. In particular, TCPFormer \citep{liu2025tcpformer}, FinePOSE \citep{Xu_2024_CVPR_finepose}, and MotionAGFormer \citep{motionagformer2024} all train on Human3.6M \citep{h36m} and test only on 3DPW \citep{3dpw}. Although each introduces novelty—TCPFormer in lifting, FinePOSE in architectural design, and MotionBERT and MotionAGFormer in binary decisions regarding the nature of the test image—none have demonstrated the ability to generalize across multiple datasets, let alone real-world scenarios. In contrast, PoseBench3D automates cross-dataset evaluation, eliminating long-existing friction in benchmarking across datasets and addressing a key limitation that has held back progress in the field.

\textbf{Shortage of Testing Environments.}
Although we commend AdaptPose \cite{Gholami_2022_CVPR} for evaluating its fine-tuned version of \cite{Pavllo_2019_CVPR} and \cite{2d3d_gong2021poseaug} across four datasets—Human3.6M, 3DHP, 3DPW, and Ski‑Pose—they do not provide a standardized evaluation framework for others to test their own models. Wang \etal \cite{i3d_cd_wang2020predicting}, Manzur \emph{et al.} \cite{manzur2024human}, and Gong \emph{et al.} \cite{2d3d_gong2021poseaug} also report cross-dataset results, which aligns more closely with our goals. Nevertheless, there remains a lack of standardized cross-dataset evaluation frameworks that enable consistent benchmarking across multiple works.

\section{Overview of PoseBench3D}

\begin{figure}[!htb]
    \centering
    \includegraphics[width=1\linewidth]{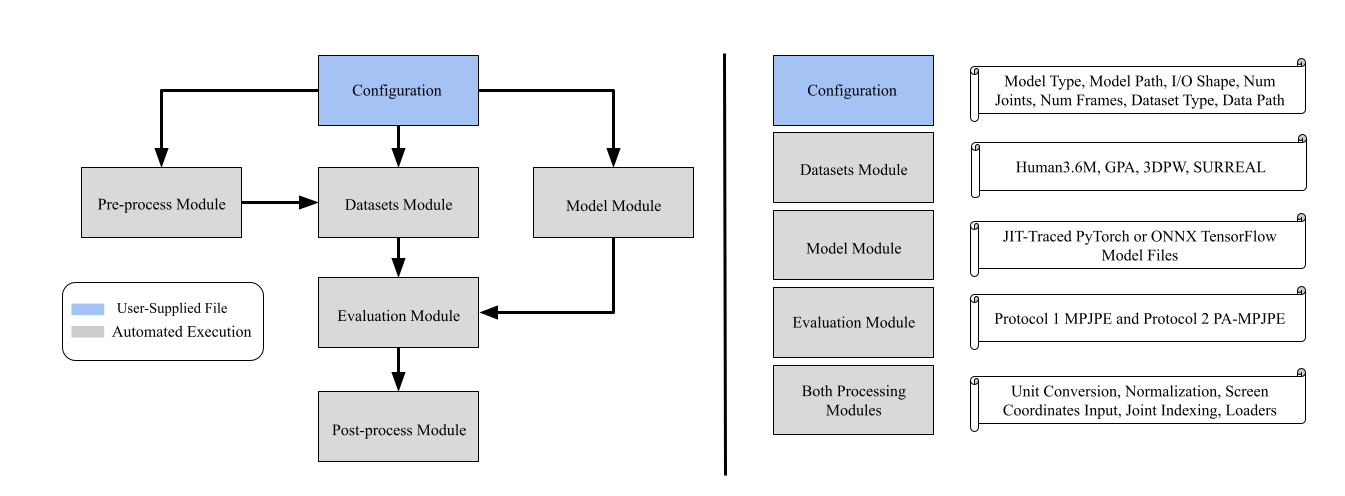}
    \caption{\centering High-level overview of interactions among the framework components.}
    \label{fig:cross-dataset-overview}
\end{figure}

Our framework, illustrated in the high-level overview in \Cref{fig:cross-dataset-overview}, is designed to support the evaluation of 2D-to-3D human pose lifting networks. In what follows, we provide a brief summary of the supported 2D-to-3D pose estimation datasets, outline the curated set of benchmark models included in our evaluation, and describe the techniques employed to standardize these datasets under a unified and fair comparison framework.

\vspace{-1em}

\subsection{Benchmark Datasets}
\label{ssec:datasets_overview}

Our framework currently supports four datasets for evaluation, depicted in \cref{tab:dataset_comparison}. The Human3.6M dataset (H36M) \cite{h36m} consists of over 3.6 million annotated 3D human poses from 15 natural activity scenarios. This dataset is currently the most popular for use in the field of 3D human pose estimation, but its controlled lab environment limits its generalization potential. Geometric Pose Affordance (GPA) \cite{gpa} focuses on the interaction between humans and objects. With $\approx$ 305,000 RGB images, GPA emphasizes environmental interaction. The 3D Poses in the Wild (3DPW) dataset \citep{3dpw} emphasizes real-world, unconstrained environments with pose data captured using IMU sensors along with automatically generated 2D-to-3D pose associations. SURREAL \citep{varol17_surreal} is a large synthetic dataset for 3D pose estimation, derived from the CMU Motion Capture database \citep{cmumocap}. It uses MoSh \cite{mosh} to match SMPL \cite{smpl} parameters to raw marker paths. Since the dataset is synthetic and uses human skeletal models, both 2D and 3D ground truth joint locations are available to arbitrary precision. 

\begin{table*}[!t]
    \centering
    \caption{Comparative summary of four benchmark 3D human pose datasets, highlighting differences in camera systems, data modalities, and geometric attributes. Each dataset is color coded through out the paper for ease of comparison.}
    \label{tab:dataset_comparison}
    \vspace{0.5em}
    \scriptsize
    \renewcommand{\arraystretch}{1.3}
    \begin{tabularx}{\textwidth}
      {>{\raggedright\arraybackslash}p{2.6cm}
       >{\raggedright\arraybackslash}X
       >{\raggedright\arraybackslash}X
       >{\raggedright\arraybackslash}X
       >{\raggedright\arraybackslash}X}
        \toprule
        & \multicolumn{4}{c}{\textbf{Datasets}} \\   
        \cmidrule(lr){2-5}
         \textbf{Criteria} & \textbf{\textcolor[HTML]{1F77B4}{Human3.6M}} & \textbf{\textcolor[HTML]{2CA02C}{GPA}} & \textbf{\textcolor[HTML]{FF7F0E}{3DPW}} & \textbf{\textcolor[HTML]{D62728}{SURREAL}} \\
        \midrule
        No. of Cameras & 4 RGB + 1 TOF & 2 RGB + 3 RGBD & 1 Moving Camera & 1 Virtual Camera \\\hline
        Subjects \& Activities & 11 Actors, 15 Daily Actions & 13 Subjects, Scripted Interactions & 7 Subjects, 18 Clothing Types & Synthetic, 6.5 M Frames from CMU MoCap \\\hline
        Environment & Indoor Lab & Indoor Studio, Rich 3D Scenes & Outdoor Real-World Settings & Synthetic Indoor Scenes \\\hline
        Imaging Space & $1000\times1002$ & $1920\times1080$ & $1920\times1080$ & $320\times240$ \\\hline
        Camera Distance (m) & $5.2 \pm 0.8$ & $5.1 \pm 1.2$ & $3.5 \pm 0.7$ & $8.0 \pm 1.0$ \\\hline
        Camera Height (m) & $1.6 \pm 0.05$ & $1.0 \pm 0.3$ & $0.6 \pm 0.8$ & $0.9 \pm 0.1$ \\\hline
        Focal Length (mm) & $1146.8 \pm 2.0$ & $1172.4 \pm 121.3$ & $1962.2 \pm 1.5$ & $600 \pm 0$ \\\hline
        No. of Joints & 38 & 34 & 24 & 24 \\
        \bottomrule
    \end{tabularx}
\end{table*}

\subsection{Supported Model Architectures}
\label{ssec:model_architectures}
Our framework is set-up to support Pose Lifting Networks models out-of-the-box, with the goal of supporting video- and image-based models in future work. 2D-to-3D pose estimation models, a.k.a. ``Lifting Networks'', are widely used for their flexibility, augmentation capabilities, and ease of implementation. Given that 2D keypoint-based representations lack many visual cues present in images (such as occlusion), it simplifies the task of 3D pose estimation–often at the cost of generalization \cite{i3d_cd_wang2020predicting, manzur2024human}. Therefore, we survey the lifting networks reported by previous works \cite{2d3d_martinez2017simple, 2d3d_gcn_zhao2019semantic, zhao2022graformer, yu2023gla, zhang2022mixste, st-gcn, diffusion_feng2023diffpose, 2d3d_gong2021poseaug, zheng2021poseformerv1, zhao2023poseformerv2, 2d3d_gcn_zeng2020srnet, 2d3d_gcn_ci2019optimizing, 2d3d_gcn_zou2021modulated} and pick 18 configurations based on their availability and adaptability. Gong \emph{et al.} \cite{2d3d_gong2021poseaug} reported results on Martinez \emph{et al.} \cite{2d3d_martinez2017simple}, Zhao \emph{et al.} \cite{2d3d_gcn_zhao2019semantic}, Pavllo \emph{et al.} \cite{vid_pavllo20193d}, and Cai \emph{et al.} \cite{st-gcn} with optimized and un-optimized variants tested on the H36M and 3DPW datasets. Since the checkpoints are already provided, we used these checkpoints to re-evaluate the models on all four datasets with our framework. We also gathered five transformer-based models \cite{zheng2021poseformerv1, zhao2023poseformerv2, zhao2022graformer, peng2024ktpformer, li2022mhformer}, two graph convolution models \cite{yu2023gla, 2d3d_gcn_zhao2019semantic}, a spatio-temporal encoder model \cite{zhang2022mixste}, and a diffusion-based model \cite{diff_hypo_shan2023diffusionhpe}. Given that some of these works omitted some checkpoints or were trained on 17 joint skeleton configurations, we decided to standardize testing by retraining these models from scratch on the more common 16 joint skeleton configuration to remain fair and consistent across comparisons. We further ensure the retrained model is as good as the original one by comparing the reported same-dataset results directly.

\subsection{Dataset Preparation}
\label{ssec:datapreparation}
To unify four distinct datasets under a common processing interface, we first identified the core elements required by lifting-based 3D pose estimation networks. These models typically take 2D keypoints as input and predict 3D joint coordinates as output. The 3D pose is usually expressed either in world-space or camera-space, with the latter being both statistically justified and widely adopted due to its alignment with image-space supervision. Additionally, most models standardize the pose by centering the skeleton at the hip joint. As shown in \cref{tab:dataset_comparison} and discussed in \citep{i3d_cd_wang2020predicting}, datasets differ significantly in terms of joint definitions, camera intrinsics/extrinsics, image resolution, and other parameters. As such, each dataset requires tailored preprocessing pipelines. For efficiency, we preprocessed and cached all dataset files as NumPy zipped archives, enabling faster data loading. Although different datasets call for different preprocessing techniques, the most important operation is acquiring and converting 3D joint positions into camera coordinates. Once this is established, projecting to 2D space follows a standardized procedure.

The H36M \citep{h36m} dataset provides four cameras with their intrinsic and extrinsic matrices. While the intrinsic matrices stay the same for all 15 action sequences across 11 subjects, the extrinsic parameters (\eg, orientation and translation) change. There are 15 actions, 2 subactions per action, and 4 cameras, all acted by 11 subjects. Since the 3D pose estimation task focuses training on subjects 1, 5, 6, 7, and 8 and testing on subjects 9 and 11, we only provide data points for these subjects. To avoid redundancy, we only sample frames that have noticeable change in the pose. The world-space 3D coordinates ($\mathrm{X} \in \mathbb{R}^{N\times3}$) are then converted into camera-space coordinates with the help of $(\mathrm{R} \cdot (\mathrm{X}^\top - \mathrm{t}))^\top$, where $\mathrm{R} \in \mathbb{R}^{3\times3}$ is the rotation matrix and $\mathrm{t}$ is the translation vector. The 2D points were projected using the perspective projection equation to obtain the ground-truth 2D keypoints. On the other hand, the GPA \citep{gpa} dataset provides intrinsic and extrinsic parameters for every frame where all the annotations are provided in a json file. The key difference in processing the dataset was the camera-space coordinate transformation. We use $\mathrm{R_i}^\top \cdot \mathrm{X}_i^\top + \mathrm{t_i}$, where the notification follows a similar convention to that of the H36M dataset. The translation vector provided was in centimeters in the original dataset, where the 3D coordinates were found to be in millimeters. The rotation matrix was also stored in a vector format. The 3DPW dataset \citep{3dpw} immediately differentiates itself from the rest of the datasets in terms of formatting. The dataset provides camera extrinsics and intrinsics per sequence, and each sequence can contain multiple subjects. We process each subject separately. Since the camera parameters are provided as an extrinsic matrix ($\mathbb{R}^{4\times4}$), a multiplication with the world space coordinates expressed as homogenous coordinates is enough to obtain the camera space coordinates. The SURREAL dataset \citep{varol17_surreal} required minimal preprocessing for camera coordinate conversion. However, we identified several invalid data batches, which consistently caused large pose estimation errors. To maintain consistency in evaluation, we excluded these anomalous samples, which accounted for less than 1\% of the dataset.

\subsection{Framework Initialization}
\label{ssec:frameworkinitiation}
Our framework consists of five key modules, as illustrated in \cref{fig:cross-dataset-overview}. Experiments are orchestrated through a global configuration file in YAML format, specifying critical parameters such as \lstinline{model_type}, \lstinline{num_workers}, \lstinline{trained_on_normalized_data}, \lstinline{output_3d}, \lstinline{video_mode}, \lstinline{num_joints}, and \lstinline{num_frames}, among others. The framework supports model checkpoints saved in either JIT or ONNX format. The \lstinline{trained_on_normalized_data} flag indicates whether the model was trained on normalized data, while the \lstinline{video_mode} toggle enables inference on multiple frames simultaneously, which is essential for temporal models. By adopting this configuration-driven approach, models can be seamlessly loaded as abstract entities, simplifying the complexity associated with conducting diverse experiments. Specifically, the Model Module initializes the chosen model, and the Dataset Module processes and prepares the datasets according to the configuration settings. The selection of joints used by the Dataset Module is determined by the \lstinline{num_joints} parameter provided in the configuration. To maintain consistency across experiments, we primarily employ a standardized 16-joint ordering: hip, right hip, right knee, right ankle, left hip, left knee, left ankle, spine, neck, head, left shoulder, left elbow, left wrist, right shoulder, right elbow, and right wrist. For certain experiments, we alternatively employ a 14-joint subset, excluding the head and spine. All evaluated models require either screen-space normalization or z-score standardization. Screen-space normalization scales pixel coordinates from the original image dimensions $(0, w)$ (width) and $(0, h)$ (height) to a unit interval $(0, 1)$. In contrast, z-score standardization involves normalizing joint coordinates, $X$ with $\frac{X - \mu}{\sigma}$, where $\mu$ is the mean and $\sigma$ is the standard deviation.

\section{Experiments and Results}
\subsection{Setup}
\label{ssec:exp_setup}

Not all datasets agree on a common seventeenth joint. Therefore, we retrain models originally trained with a 17-joint configuration using only 16 joints. Following established conventions in the field \cite{2d3d_chen20173d,2d3d_martinez2017simple,2d3d_gcn_zhao2019semantic,2d3d_chen2019unsupervised,2d3d_gcn_ci2019optimizing}, we report both Protocol 1 error—MPJPE (Mean Per-Joint Position Error)—and Protocol 2 error—PA-MPJPE (MPJPE after Procrustes Alignment), commonly referred to as Protocol 1 and Protocol 2, respectively. Since our experiments focus on lifting networks, they are not memory intensive. All experiments were conducted using two A30 24GB GPUs. Note that our framework supports both GPU and CPU systems.

\subsection{Cross-Dataset Evaluation}
We report the MPJPE scores in \cref{tab:cd_h36m_train_mpjpe} and PA-MPJPE scores in \cref{tab:cd_h36m_train_pampjpe}. We also include two cross-dataset results available from prior work marked with $\ddagger$ in the tables.
 
Most methods generalize only partially, often suffering severe performance drops. Classical graph-based models such as SEM-GCN \citep{2d3d_gcn_zhao2019semantic} and ST-GCN \citep{st-gcn}, as well as recent transformer models like PoseFormer V1/V2 \citep{zheng2021poseformerv1, zhao2023poseformerv2} and GraFormer \citep{zhao2022graformer}, exhibit large errors on GPA and 3DPW—frequently exceeding 200 mm and occasionally 300 mm. PoseAug-optimized variants (\eg, Martinez$\dagger$, ST-GCN$\dagger$, VideoPose$\dagger$) improve over their non-optimized counterparts. Even transformer architectures with excellent same-dataset (H36M) scores overfit sharply, performing poorly on synthetic (SURREAL) and in-the-wild (3DPW) data. This suggests that such models may lack the necessary inductive biases or training regularization to generalize across data domains. Among all evaluated models, Manzur \emph{et al.} \citep{manzur2024human} consistently records the lowest MPJPE on every dataset. Wang et al. \cite{i3d_cd_wang2020predicting} comes in second. Both of these methods take relative viewpoint into account, a strong indication of the significance of viewpoint in 3D HPE.


\begin{table*}[!t]
  \centering
  \caption{Cross-dataset evaluation sorted by decreasing average MPJPE (mm). All models are trained on H36M. Lower is better. $\dagger$: PoseAug-optimized; $\diamond$: unoptimized variant; $\bullet$: retrained from scratch; $\ddagger$: reported from prior work.}
  \label{tab:cd_h36m_train_mpjpe}
  \small   
  \begin{tabularx}{\linewidth}{p{3.5cm} *{5}{Y}}
    \toprule
    & \multicolumn{4}{c}{\textbf{Cross-Dataset Evaluation}} \\
    \cmidrule(lr){2-5}
    \textbf{Model Name} & \textbf{\textcolor[HTML]{1F77B4}{Human3.6M}} & \textbf{\textcolor[HTML]{2CA02C}{GPA}} & \textbf{\textcolor[HTML]{FF7F0E}{3DPW}} & \textbf{\textcolor[HTML]{D62728}{SURREAL}} & \textbf{Average Error (MPJPE \(\downarrow\))} \\
    \midrule
    GraFormer \citep{zhao2022graformer} $\bullet$ & {36.44} & 259.11 & 308.96 & 150.46 & 188.74 \\
    SEM-GCN \citep{2d3d_gong2021poseaug} $\diamond$ & 47.03 & 262.34 & 315.31 & 118.30 & 185.75 \\
    SEM-GCN \citep{2d3d_gong2021poseaug} $\dagger$ & 41.90 & 241.21 & 239.07 & 107.26 & 157.36 \\
    VideoPose \citep{2d3d_gong2021poseaug} $\diamond$ & 41.47 & 208.45 & 257.81 & 107.96 & 153.92 \\
    GLA-GCN \citep{yu2023gla} $\bullet$ & 44.51 & 237.29 & 207.59 & 119.08 & 152.12 \\
    ST-GCN \citep{2d3d_gong2021poseaug} $\diamond$ & 41.52 & 205.76 & 238.48 & 107.61 & 148.34 \\
    Martinez \emph{et al.} \citep{2d3d_gong2021poseaug} $\diamond$ & 41.42 & 205.62 & 226.20 & 110.01 & 145.81 \\
    PoseFormer V1 \citep{zheng2021poseformerv1} $\bullet$ & 42.82 & 217.90 & 161.97 & 156.59 & 144.82 \\
    PoseFormer V2 \citep{zhao2023poseformerv2} $\bullet$ & 42.80 & 209.90 & 162.45 & 146.39 & 140.39 \\
    KTPFormer \citep{peng2024ktpformer} $\bullet$ & 38.12 & 205.71 & 193.63 & 108.95 & 136.60 \\
    MixSTE \citep{zhang2022mixste} $\bullet$ & 38.44 & 182.13 & 171.28 & 131.23 & 130.77 \\
    DDHPose \citep{cai2025disentangleddiffusionbased3dhuman} $\bullet$ & 38.28 & 200.29 & 138.64 & 129.15 & 126.59 \\
    D3DP \citep{diff_hypo_shan2023diffusionhpe} $\bullet$ & 39.61 & 189.74 & 148.56 & 127.90 & 126.45 \\
    ST-GCN \citep{2d3d_gong2021poseaug} $\dagger$ & 36.83 & 185.63 & 174.16 & 101.95 & 124.64 \\
    MHFormer \citep{li2022mhformer} $\bullet$ & 42.60 & 202.59 & 202.59 & 114.55 & 120.86 \\
    Martinez \emph{et al.} \citep{2d3d_gong2021poseaug} $\dagger$ & 39.11 & 169.79 & 134.12 & {98.99} & 110.50 \\
    VideoPose \citep{2d3d_gong2021poseaug} $\dagger$ & 39.02 & 174.39 & 126.05 & 100.42 & 109.97 \\
    Wang \emph{et al.} \citep{i3d_cd_wang2020predicting} $\ddagger$ & 52.00 & {98.30} & {109.5} & 114.00 & {93.45} \\
    Manzur \emph{et al.} \citep{manzur2024human} $\ddagger$ & {33.52} & {92.31} & {95.83} & {65.62} & {71.82} \\

    \bottomrule
  \end{tabularx}
\end{table*}

All of the scores reported in Protocol 1 (MPJPE) are expected to improve with Procrustes alignment, since this method involves reducing the distance between two sets of 3D points via a rigid transformation. This alignment reduces any orientation-related error, providing an interesting perspective on how purely geometric transformations affect generalization. \Cref{tab:cd_h36m_train_pampjpe} shows the scores for the same models after Procrustes alignment. The greatest improvement is observed in challenging datasets such as GPA and 3DPW, where Manzur \emph{et al.} \cite{manzur2024human} improve from $92.31$mm to $69.48$mm on GPA and from $95.83$mm to $64.28$mm on 3DPW. Even models that showed poor generalization achieve significant gains. For instance, on 3DPW, the unoptimized variants of ST-GCN$\diamond$ improve from $238.48$ to $206.45$\,mm and SEM-GCN$\diamond$ from $315.31$ to $166.76$\,mm. This outcome suggests that many models produce poses with strong internal structure but poor absolute orientation. Transformer-based architectures (\eg, PoseFormer V1/V2$\bullet$ and GraFormer$\bullet$) also attain substantial reductions in error.

\begin{table*}[!t]
  \centering
  \caption{Cross-dataset evaluation sorted by decreasing average PA-MPJPE (mm). All models are trained on H36M. Lower is better. $\dagger$: PoseAug-optimized; $\diamond$: unoptimized variant; $\bullet$: retrained from scratch; $\ddagger$: reported from prior work.}
  \label{tab:cd_h36m_train_pampjpe}
  \small

  \begin{tabularx}{\linewidth}{p{3.5cm} *{5}{Y}}

    \toprule
    & \multicolumn{4}{c }{\textbf{Cross-Dataset Evaluation}} \\
    \cmidrule(lr){2-5}
    \textbf{Model Name} &  \textbf{\textcolor[HTML]{1F77B4}{Human3.6M}} & \textbf{\textcolor[HTML]{2CA02C}{GPA}} & \textbf{\textcolor[HTML]{FF7F0E}{3DPW}} & \textbf{\textcolor[HTML]{D62728}{SURREAL}} & \textbf{Average Error (PA-MPJPE \(\downarrow\))} \\
    \midrule
    SEM-GCN \citep{2d3d_gong2021poseaug} $\diamond$ & 36.12 & 178.43 & 166.76 & 87.35 & 117.17\\
    GraFormer \citep{zhao2022graformer} $\bullet$   & {28.40} & 152.31 & 189.30 & 87.87 & 114.47\\
    ST-GCN  \citep{2d3d_gong2021poseaug} $\diamond$ & 32.47 & 125.99 & 206.45 & 69.11 & 108.51\\
    SEM-GCN \citep{2d3d_gong2021poseaug} $\dagger$  & 33.66 & 166.88 & 131.38 & 80.98 & 103.23\\
    PoseFormer V1 \citep{zheng2021poseformerv1} $\bullet$ & 33.50 & 138.12 & 103.89 & 95.72 & 92.81\\
    GLA-GCN \citep{yu2023gla} $\bullet$            & 35.27 & 148.26 & 106.10 & 73.31 & 90.74\\
    PoseFormer V2 \citep{zhao2023poseformerv2} $\bullet$ & 33.18 & 145.94 & 92.47 & 91.02 & 90.65\\
    KTPFormer \citep{peng2024ktpformer} $\bullet$  & 30.27 & 133.73 & 127.09 & 67.66 & 89.69\\
    ST-GCN  \citep{2d3d_gong2021poseaug} $\dagger$  & {28.69} & 112.12 & 131.99 & 65.17 & 84.49\\
    Martinez \citep{2d3d_gong2021poseaug} $\diamond$ & 31.80 & 124.17 & 111.30 & 67.97 & 83.81\\
    DDHPose \citep{cai2025disentangleddiffusionbased3dhuman} $\bullet$ & 30.13 & 139.24 & 76.85 & 80.95 & 81.79 \\
    VideoPose \citep{2d3d_gong2021poseaug} $\diamond$ & 32.17 & 126.61 & 102.65 & 65.67 & 81.78\\
    MixSTE \citep{zhang2022mixste} $\bullet$       & 31.05 & 120.85 & 91.12 & 76.96 & 80.00\\
    D3DP \citep{diff_hypo_shan2023diffusionhpe} $\bullet$ & 31.25 & 133.82 & {73.86} & 78.83 & 79.44\\
    MHFormer \citep{li2022mhformer} $\bullet$      & 32.56 & 124.50 & 124.50 & 69.18 & 73.21\\
    Martinez \citep{2d3d_gong2021poseaug} $\dagger$ & 30.31 & {103.26}& 79.74 & 59.94 & 68.31\\
    VideoPose \citep{2d3d_gong2021poseaug} $\dagger$ & 30.17 & 108.92 & 75.29 & {58.33} & {68.18}\\
    Manzur \emph{et al.} \citep{manzur2024human} $\ddagger$       & 29.10 & {69.48} & {64.28} & {51.53} & {53.60}\\
    \bottomrule
  \end{tabularx}
\end{table*}

\begin{figure}[!htb]
    \centering
    \includegraphics[width=1\linewidth]{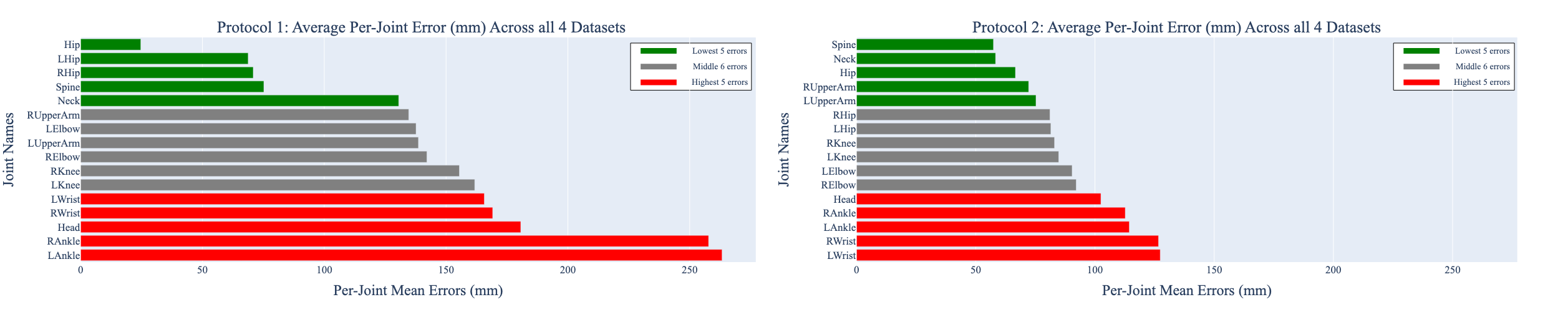}
    \caption{\centering Per-joint errors for Protocols 1 and 2, averaged across all models and datasets.}
    \label{fig:joint_errors}
\end{figure}

We also analyze per-joint position errors under both protocols in \cref{fig:joint_errors}, averaged across all 18 model configurations and all four datasets. While the endpoints of the human body—such as the wrists, ankles, and head—are more susceptible to errors, note that Procrustes alignment alone results in a substantial reduction in these errors (highlighted in red). Since Procrustes alignment accounts for rigid transformations, this highlights that human “pose” estimation is distinct from “position” estimation. Therefore, future efforts should focus on reducing angular errors in the bones rather than solely minimizing positional discrepancies.

\subsection{Impact of Z-Score Standardization}

Z-score standardization is a common operation to normalize the data by zero-centering with the mean and scaling by the standard deviation. In the context of pose estimation, where datasets may originate from diverse distributions, such normalization indeed improves cross-dataset performance when the mean and standard deviation are known—though they are typically unavailable during inference time in real-world scenarios. \Cref{tab:mpjpe_martinez_semgcn_normalization} shows the models \cite{2d3d_martinez2017simple} and \cite{2d3d_gcn_zhao2019semantic} trained with H36M and normalized with the mean and standard deviation of the same dataset. For the other datasets, we used their respective mean and standard deviation to normalize them. We also include the unoptimized ($\diamond$) and PoseAug-optimized ($\dagger$) variants from \cref{tab:cd_h36m_train_mpjpe} for comparison, and also note the percentage improvement relative to the unoptimized variant in both cases. From all the cases, note that Z-score normalization performed with the test set's mean and standard deviation improves cross-dataset generalization significantly for both models (\eg, SemGCN sees an improvement of more than 50\% in 3DPW). PoseAug's optimization adds novel poses to the training and secures a place right in the middle of the unoptimized variant and the Z-score standardized variant. This shows that models overfit on the distribution they are trained on—which can often be countered by increasing variation.


\begin{table*}[!t]  
  \centering
 \caption{MPJPE comparison of Martinez and SEM-GCN models across datasets. Percentage improvements relative to the baseline ($\diamond$) are in parentheses. Arrows in green indicate improvement over the Unoptimized (baseline), and red arrows indicate degradation from baseline.}
\label{tab:mpjpe_martinez_semgcn_normalization}

\begin{subtable}{\linewidth}
  \centering
  \caption{Martinez Comparison}
  \begingroup
    \setlength{\tabcolsep}{3pt}
    \footnotesize
        \begin{tabularx}{\linewidth}{>{\raggedright\arraybackslash}X@{} *{5}{Y}}

      \toprule
      \textbf{Model} & \textbf{H36M} & \textbf{GPA} & \textbf{3DPW} & \textbf{SURREAL} & \textbf{Average Error} \\
      \midrule
      Unoptimized $\diamond$  & 41.42 & 205.62 & 226.20 & 110.01 & 145.81 \\
      Optimized $\dagger$ & 39.11 (\textcolor{green}{$\downarrow$} 5.6\%) & 169.79 (\textcolor{green}{$\downarrow$} 17.4\%) & 134.12 (\textcolor{green}{$\downarrow$} 40.7\%) & 98.99 (\textcolor{green}{$\downarrow$} 10.0\%) & 110.50 (\textcolor{green}{$\downarrow$} 24.2\%) \\
      Z-score Normalization            & 52.37 (\textcolor{magenta}{$\uparrow$} 26.4\%) & 104.39 (\textcolor{green}{$\downarrow$} 49.3\%) & 141.10 (\textcolor{green}{$\downarrow$} 37.6\%) & 81.64 (\textcolor{green}{$\downarrow$} 25.8\%) & 94.88 (\textcolor{green}{$\downarrow$} 34.9\%) \\
      \bottomrule
    \end{tabularx}
  \endgroup
\end{subtable}

\vspace{0.8em} 

\begin{subtable}{\linewidth}
  \centering
  \caption{SEM-GCN Comparison}
  \begingroup
    \setlength{\tabcolsep}{3pt}
    \footnotesize
    \begin{tabularx}{\linewidth}{>{\raggedright\arraybackslash}X@{} *{5}{Y}}

      \toprule
      \textbf{Model} & \textbf{H36M} & \textbf{GPA} & \textbf{3DPW} & \textbf{SURREAL} & \textbf{Average Error} \\
      \midrule
      Unoptimized $\diamond$  & 47.03 & 262.34 & 315.31 & 118.30 & 185.75 \\
      Optimized $\dagger$ & 41.90 (\textcolor{green}{$\downarrow$} 10.9\%) & 241.21 (\textcolor{green}{$\downarrow$} 8.1\%) & 239.07 (\textcolor{green}{$\downarrow$} 24.2\%) & 107.26 (\textcolor{green}{$\downarrow$} 9.3\%) & 157.36 (\textcolor{green}{$\downarrow$} 15.3\%) \\
      Z-score Normalization           & 53.94 (\textcolor{magenta}{$\uparrow$} 14.7\%) & 114.85 (\textcolor{green}{$\downarrow$} 56.2\%) & 153.61 (\textcolor{green}{$\downarrow$} 51.3\%) & 99.88 (\textcolor{green}{$\downarrow$} 15.6\%) & 105.57 (\textcolor{green}{$\downarrow$} 43.2\%) \\
      \bottomrule
    \end{tabularx}
  \endgroup
\end{subtable}

\end{table*}

\subsection{Impact of Viewpoint}
\label{ssec:impact_of_vp}

\begin{figure}
    \centering
    \begin{subfigure}{0.4\linewidth}
        \centering
        \includegraphics[width=0.95\linewidth]{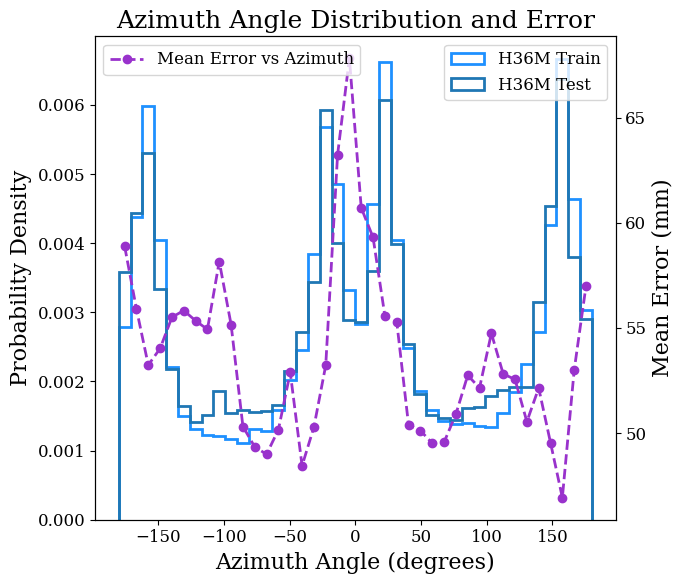}
        \caption{H36M}
    \end{subfigure}
    \begin{subfigure}{0.4\linewidth}
        \centering
        \includegraphics[width=0.95\linewidth]{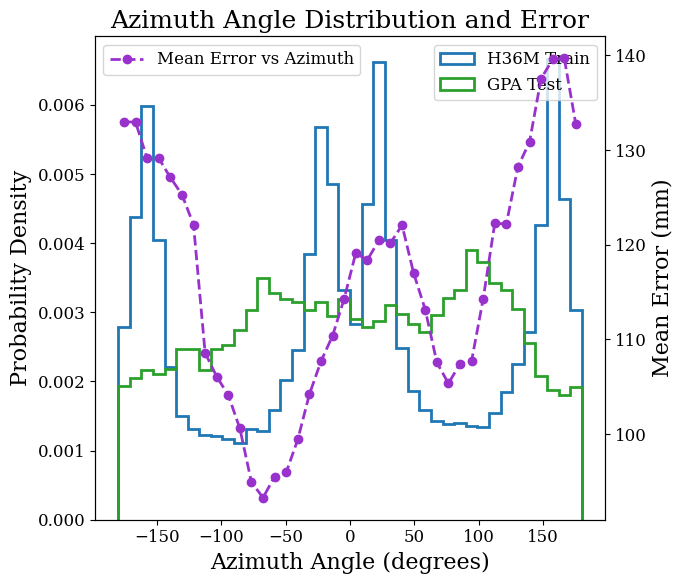}
        \caption{GPA}
    \end{subfigure}
    \begin{subfigure}{0.4\linewidth}
        \centering
        \includegraphics[width=0.95\linewidth]{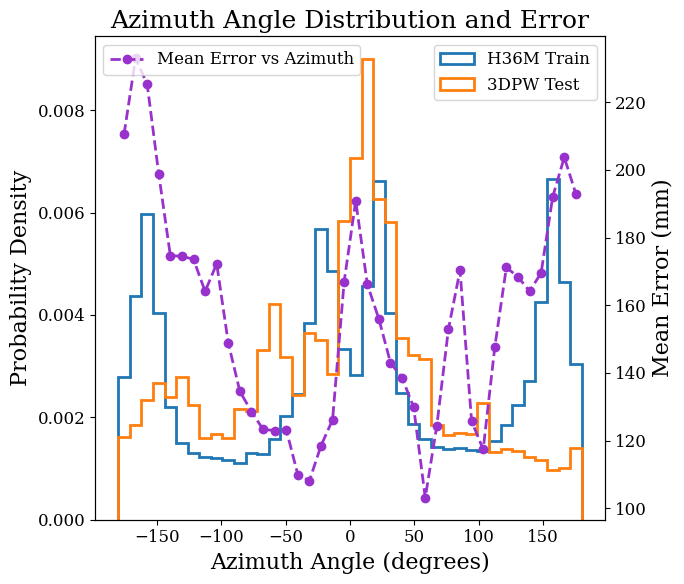}
        \caption{3DPW}
    \end{subfigure}
    \begin{subfigure}{0.4\linewidth}
        \centering
        \includegraphics[width=0.95\linewidth]{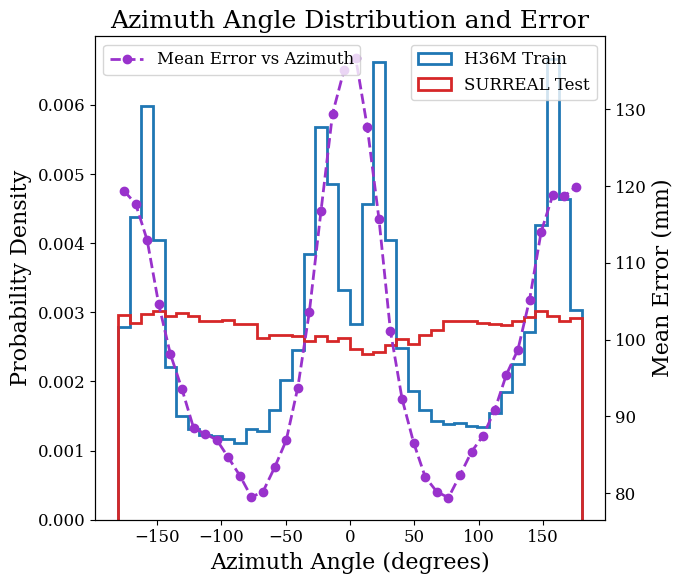}
        \caption{SURREAL}
    \end{subfigure}
    \caption{MPJPE score (SEM-GCN \cite{2d3d_gcn_zhao2019semantic} trained on H36M) compared against viewpoint distribution in azimuth relative to the subject.}
    \label{fig:azim_err_h36m_train}
\end{figure}

\begin{figure}
    \centering
    \begin{subfigure}{0.4\linewidth}
        \centering
        \includegraphics[width=0.95\linewidth]{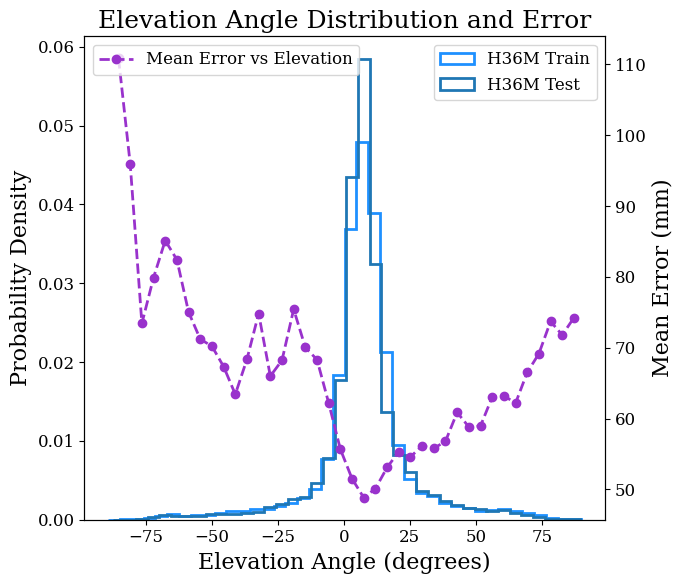}
        \caption{H36M}
    \end{subfigure}
    \begin{subfigure}{0.4\linewidth}
        \centering
        \includegraphics[width=0.95\linewidth]{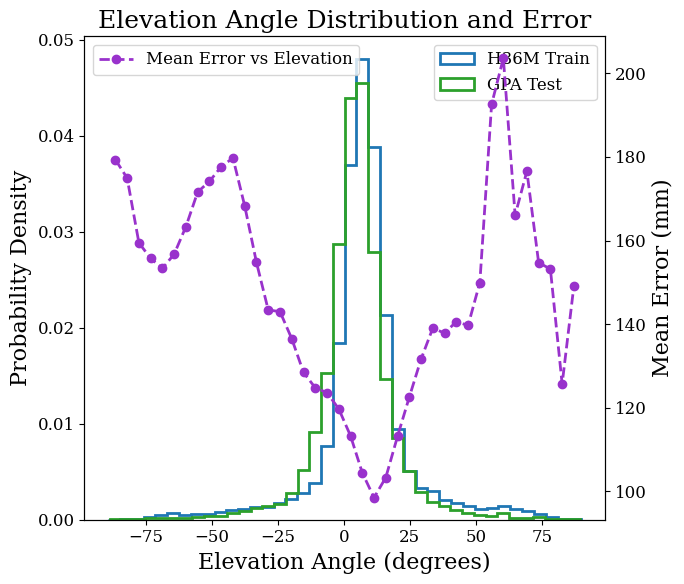}
        \caption{GPA}
    \end{subfigure}
    \begin{subfigure}{0.4\linewidth}
        \centering
        \includegraphics[width=0.95\linewidth]{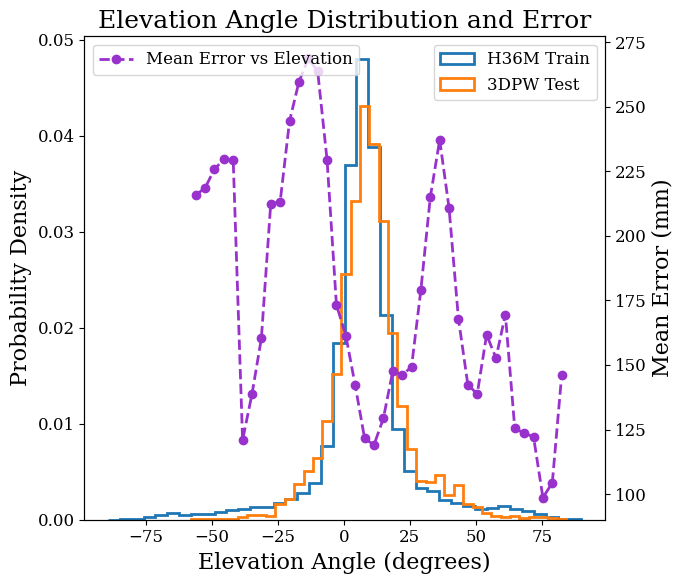}
        \caption{3DPW}
    \end{subfigure}
    \begin{subfigure}{0.4\linewidth}
        \centering
        \includegraphics[width=0.95\linewidth]{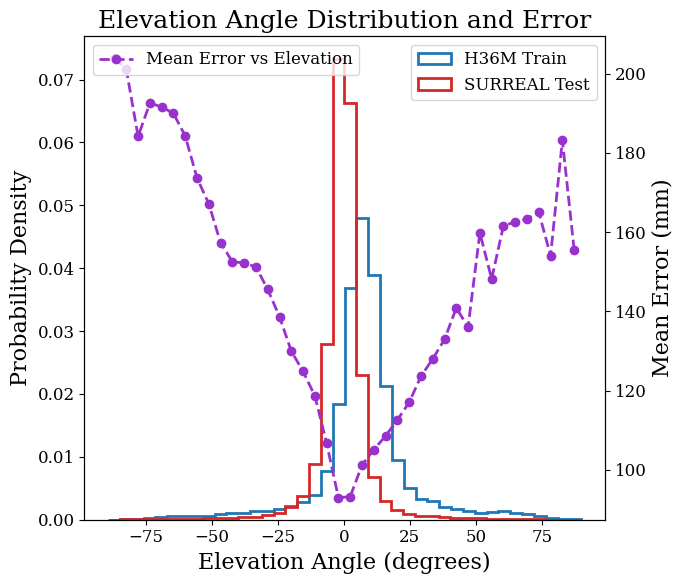}
        \caption{SURREAL}
    \end{subfigure}
    \caption{MPJPE score (SEM-GCN \cite{2d3d_gcn_zhao2019semantic} trained on H36M) compared against viewpoint distribution in elevation relative to the subject.}
    \label{fig:elev_err_h36m_train}
\end{figure}

\begin{figure}[!htb]
    \centering
    \includegraphics[width=.9\linewidth]{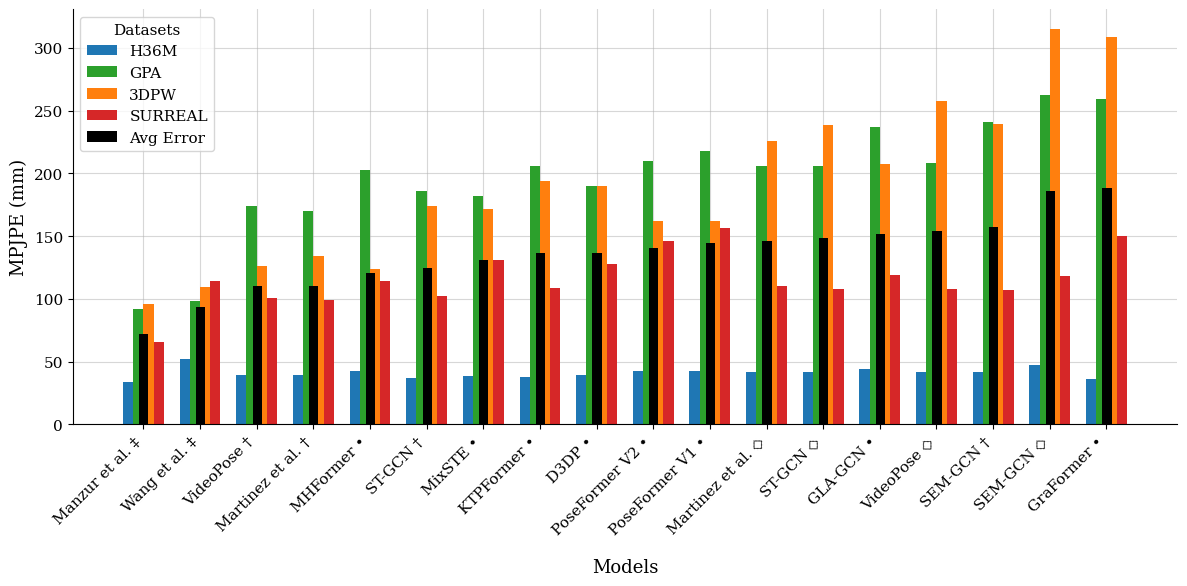}
    \caption{\centering Protocol 1 visual results from \cref{tab:cd_h36m_train_mpjpe} shown in increasing average dataset error.}
    \label{fig:extra_visualizations_1}
\end{figure}

\begin{figure}[!htb]
    \centering
    \includegraphics[width=.9\linewidth]{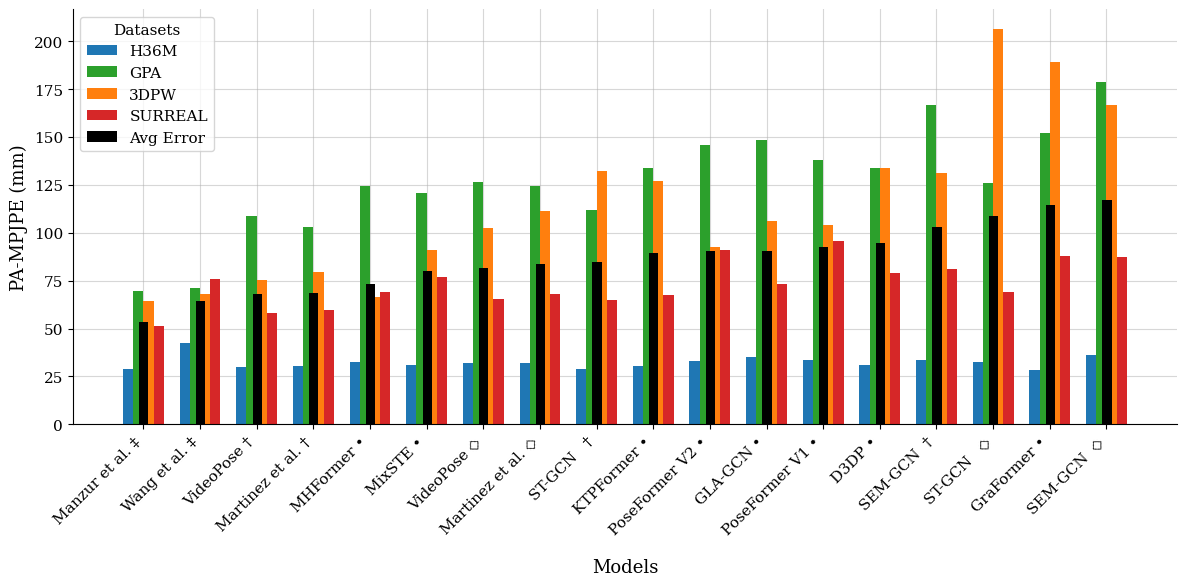}
    \caption{\centering Protocol 2 visual results from \cref{tab:cd_h36m_train_pampjpe} shown in increasing average dataset error.}
    \label{fig:extra_visualizations_2}
\end{figure}

One interesting experiment is to look at the interplay between the viewpoint distribution of the datasets and its impact on the MPJPE error. In \cref{fig:azim_err_h36m_train,fig:elev_err_h36m_train}, we present the MPJPE scores against the viewpoint distribution over the training and testing sets simultaneously. One common trend across all these figures is that whenever the viewpoint distribution disagrees with the training set's viewpoint distribution, we observe a spike in error. This shows how much critical viewpoint is in 3D human pose estimation. \Cref{tab:spearman_corr} formally measures the inverse relationship between training data frequency and joint error by computing the Spearman correlation between the two, measured across patches on the viewpoint sphere measuring \ang{5} $\times$ \ang{10} degrees in elevation and azimuth, respectively. The correlation is always negative with high statistical significance. Notice how the p-value is lower when the evaluation is cross-dataset, indicating the variance in viewpoint distribution severely impacts performance. Please refer to \Cref{fig:extra_visualizations_1} and \Cref{fig:extra_visualizations_2} for greater comparison of error rates across models and datasets. 

\begin{table}[!htb]
    \centering
    \caption{Spearman correlation (which contained at least 5 training images and at least 5 test images) between training viewpoint distributions and test error (\ang{5} elev × \ang{10} azim bins). Num = number of bins; sigma is the p-value represented as the number of standard deviations from random.}
    \label{tab:spearman_corr}
    \scriptsize
    \begin{tabular}{rllrrr}
        \toprule
        \textbf{Num} & \textbf{Train} & \textbf{Test} & \textbf{Spearman} & \textbf{P-val} & \textbf{Sigma} \\
        \midrule
        377 & 3DPW    & GPA     & -0.45 & \num{1.1e-18} & 9.78 \\
        380 & 3DPW    & H36M    & -0.30 & \num{2.4e-07} & 6.10 \\
        388 & 3DPW    & SURREAL & -0.64 & \num{1.1e-47} & 16.38 \\
        304 & 3DPW    & 3DPW    & -0.44 & \num{2.8e-14}   & 8.55 \\
        377 & SURREAL & GPA     & -0.33 & \num{2.6e-09} & 6.85 \\
        380 & SURREAL & H36M    & -0.19 & \num{0.01838}   & 3.70 \\
        388 & SURREAL & SURREAL & -0.47 & \num{1.2e-21} & 10.55 \\
        304 & SURREAL & 3DPW    & -0.35 & \num{1.5e-08} & 6.58 \\
        641 & H36M    & GPA     & -0.36 & \num{3.4e-19} & 9.78 \\
        939 & H36M    & H36M    & -0.55 & \num{1.9e-78} & 20.36 \\
        891 & H36M    & SURREAL & -0.61 & \num{7.5e-96} & 22.90 \\
        417 & H36M    & 3DPW    & -0.20 & \num{0.003701}  & 4.13 \\
        621 & GPA     & GPA     & -0.48 & \num{1.3e-36} & 13.71 \\
        738 & GPA     & H36M    & -0.63 & \num{3.0e-87} & 22.02 \\
        751 & GPA     & SURREAL & -0.68 & \num{6.8e-112} & 25.50 \\
        416 & GPA     & 3DPW    & -0.39 & \num{1.3e-14} & 8.58 \\
        \bottomrule
    \end{tabular}
\end{table}

\subsection{Correlation between Viewpoint and Error}
\label{ssec:correlation_viewpoint_error}
\begin{figure}[!htb]
    \centering
    \begin{subfigure}{0.31\linewidth}
        \centering
        \begin{subfigure}{1.0\linewidth}
            \centering
            \includegraphics[width=0.85\linewidth]{figs/Elevation_Figs/GPA_Error_Figures/ModelTrainedOnH36M/elevation_plot.png}
        \end{subfigure}
        \begin{subfigure}{1.0\linewidth}
            \centering
            \includegraphics[width=0.85\linewidth]{figs/Azimuth_Figs/GPA_Error_Figures/ModelTrainedOnH36M/azimuth_plot.png}
        \end{subfigure}
        \caption{}\label{fig:elev_azim_err_h36m_train_gpa_test}
    \end{subfigure}
    \begin{subfigure}{0.65\linewidth}
        \centering
        \includegraphics[width=0.9\linewidth]{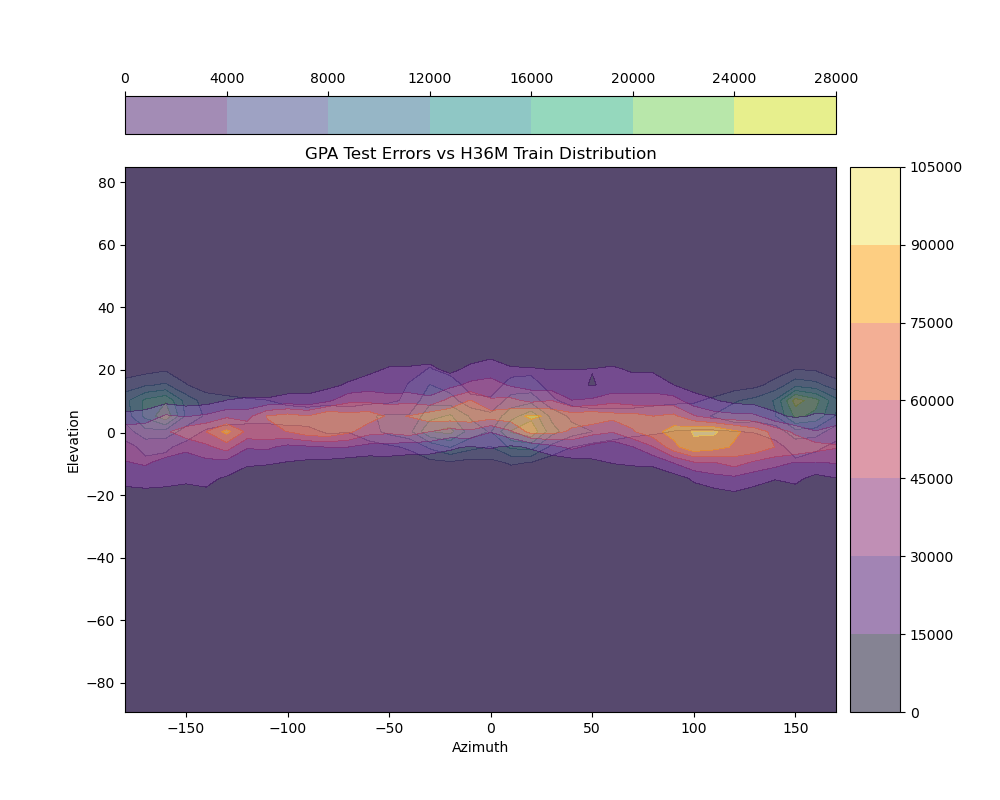}
        \caption{}\label{fig:contour_h36m_train_gpa_test}
    \end{subfigure}
    \caption{(a) Viewpoint distribution vs. MPJPE (mm) error. (b) Contour plot showing the error (in red) and viewpoint distribution (in green) with the x and y axes as azimuth and elevation, respectively. For both figures, errors are obtained on the GPA test set from \cite{2d3d_gcn_zhao2019semantic} trained with the H36M dataset.}
    \label{fig:viewpoint_err_h36m_train_gpa_test}
\end{figure}

\begin{figure}[!htb]
    \centering
    \begin{subfigure}{0.31\linewidth}
        \centering
        \begin{subfigure}{1.0\linewidth}
            \centering
            \includegraphics[width=0.85\linewidth]{figs/Elevation_Figs/3DPW_Error_Figures/ModelTrainedOnH36M/elevation_plot.png}
        \end{subfigure}
        \begin{subfigure}{1.0\linewidth}
            \centering
            \includegraphics[width=0.85\linewidth]{figs/Azimuth_Figs/3DPW_Error_Figures/ModelTrainedOnH36M/azimuth_plot.png}
        \end{subfigure}
        \caption{}\label{fig:elev_azim_err_h36m_train_3dpw_test}
    \end{subfigure}
    \begin{subfigure}{0.65\linewidth}
        \centering
        \includegraphics[width=0.9\linewidth]{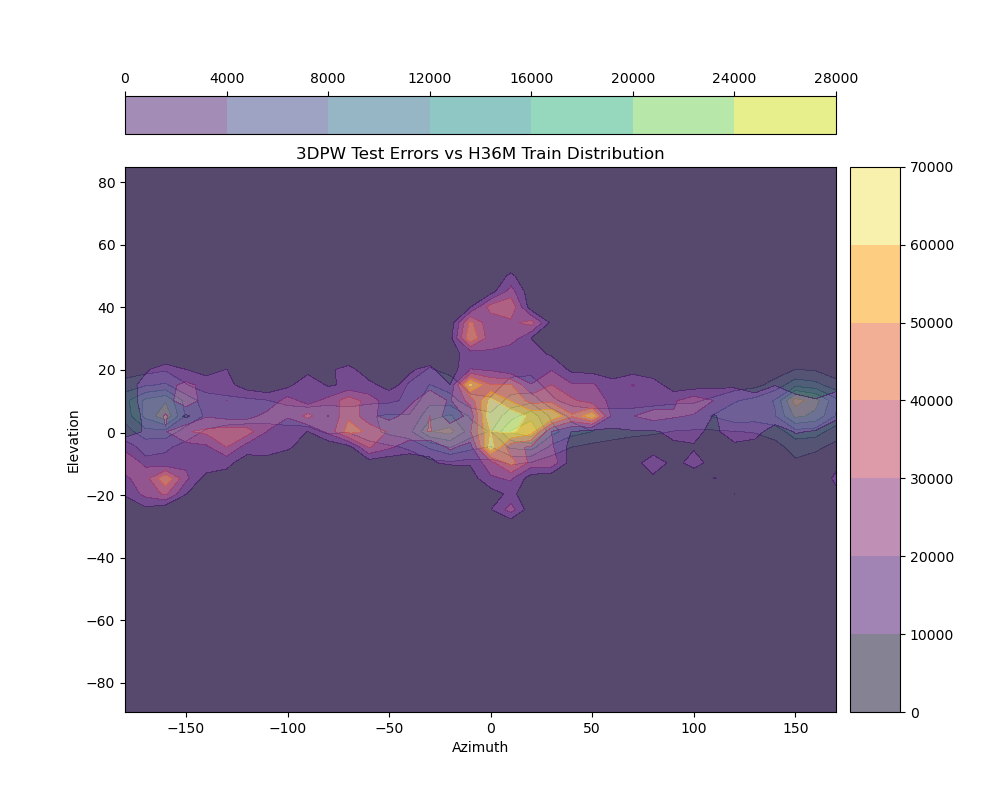}
        \caption{}\label{fig:contour_h36m_train_3dpw_test}
    \end{subfigure}
    \caption{(a) Viewpoint distribution vs. MPJPE (mm) error. (b) Contour plot showing the error (in red) and viewpoint distribution (in green) with the x and y axes as azimuth and elevation, respectively. For both figures, errors are obtained on the 3DPW test set from \cite{2d3d_gcn_zhao2019semantic} trained with the H36M dataset.}
    \label{fig:viewpoint_err_h36m_train_3dpw_test}
\end{figure}

\begin{figure}[!htb]
    \centering
    \begin{subfigure}{0.31\linewidth}
        \centering
        \begin{subfigure}{1.0\linewidth}
            \centering
            \includegraphics[width=0.85\linewidth]{figs/Elevation_Figs/Surreal_Error_Figures/ModelTrainedOnH36M/elevation_plot.png}
        \end{subfigure}
        \begin{subfigure}{1.0\linewidth}
            \centering
            \includegraphics[width=0.85\linewidth]{figs/Azimuth_Figs/Surreal_Error_Figures/ModelTrainedOnH36M/azimuth_plot.png}
        \end{subfigure}
        \caption{}\label{fig:elev_azim_err_h36m_train_surreal_test}
    \end{subfigure}
    \begin{subfigure}{0.65\linewidth}
        \centering
        \includegraphics[width=0.9\linewidth]{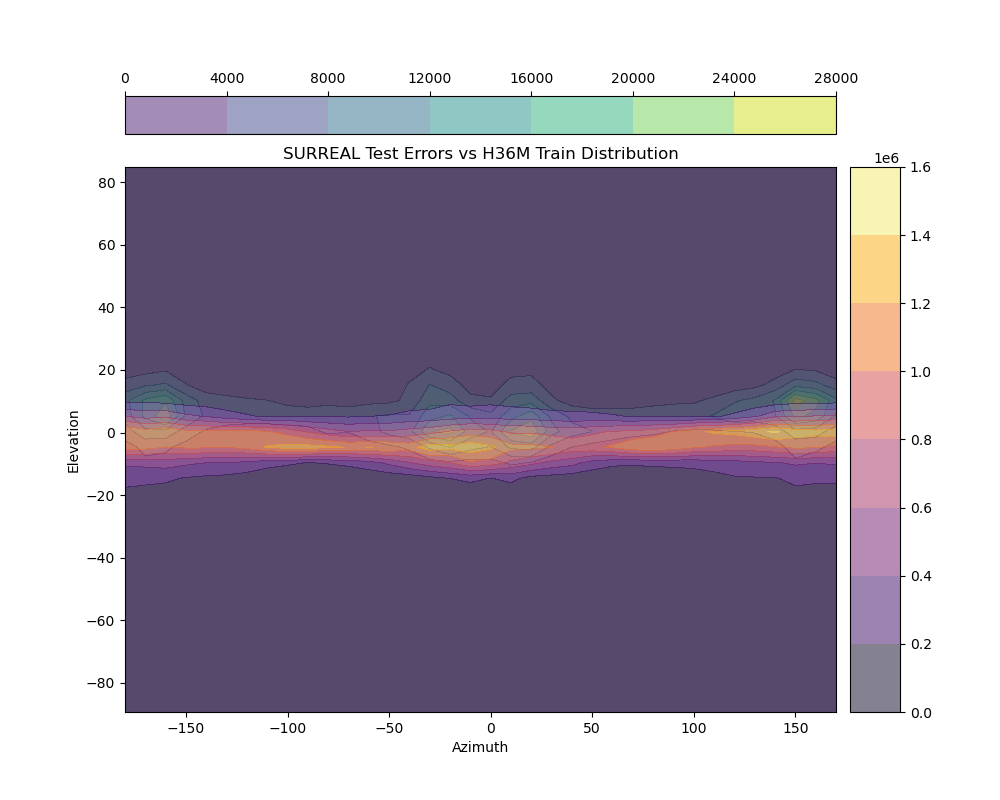}
        \caption{}\label{fig:contour_h36m_train_surreal_test}
    \end{subfigure}
    \caption{(a) Viewpoint distribution vs. MPJPE (mm) error. (b) Contour plot showing the error (in red) and viewpoint distribution (in green) with the x and y axes as azimuth and elevation, respectively. For both figures, errors are obtained on the SURREAL test set from \cite{2d3d_gcn_zhao2019semantic} trained with the H36M dataset.}
    \label{fig:viewpoint_err_h36m_train_surreal_test}
\end{figure}

In \cref{fig:azim_err_h36m_train,fig:elev_err_h36m_train}, we presented the error distribution curve with respect to the elevation and azimuth distribution separately. However, azimuth and elevation alone do not dictate the error---the number of training figures at a given point on the sphere dictates the error (\emph{i.e.}, both elevation and azimuth together). Therefore, we include additional contour plots showing the viewpoint and error distribution. \Cref{tab:spearman_corr} showed a strong inverse correlation between the error and viewpoint distribution with high statistical significance. This is not observed in individual azimuth plots (\eg, \cref{fig:azim_err_h36m_train}). In the contour plots (\eg, \cref{fig:contour_h36m_train_gpa_test,fig:contour_h36m_train_3dpw_test,fig:contour_h36m_train_surreal_test}), the viewpoint and error distribution are marked with green and red heatmaps. The X and Y axes represent azimuth and elevation, respectively. Whenever there is a trough in the viewpoint distribution---\emph{i.e.} the training samples are under-sampled---the error goes up. This shows how much critical viewpoint is in 3D human pose estimation.

\section{Conclusion and Future Work}
\label{sec:conclusion}
We have introduced a framework for evaluating cross-dataset performance of 2D-to-3D pose lifting networks. PoseBench3D enables standardized, reproducible evaluation and supports automatic benchmarking across multiple datasets. We apply PoseBench3D to evaluate 18 configurations of models, using checkpoints either obtained from prior work or retrained according to the original authors' specifications. With more than 100 newly reported cross-dataset comparisons, we analyzed the results through several lenses and identify key factors—such as viewpoint distribution and normalization strategies—that significantly impact generalization in 3D human pose estimation.

The limitation of this work is that the current framework supports only four widely used datasets and is limited to pose lifting networks. In the future, we plan to extend PoseBench3D to support image-based models, enabling evaluation of end-to-end systems that go from RGB input to 3D pose. We also aim to integrate additional datasets and enhance the modularity of the framework to further encourage community contributions and broader adoption. Ultimately, we hope PoseBench3D will serve as a foundation for fair, comprehensive, and scalable benchmarking in 3D human pose estimation.

{
    \small
    \bibliographystyle{ieeenat_fullname}
    \bibliography{main}
}


\end{document}